\documentclass[runningheads]{llncs}
\usepackage[T1]{fontenc}
\usepackage{amsmath}
\usepackage{amsfonts}
\usepackage{graphicx}
\usepackage{tcolorbox}
\newcommand{\charname}{\mathcal{C}}
\newcommand{\keywords}[1]{k(#1)}
\newcommand{\prompt}[2]{\mathcal{P}^{+}(#1,#2)}
\newcommand{\gendes}{\mathcal{GD}}
\newcommand{\chardes}{\mathcal{CD}}
\newcommand{\Pin}{\mathcal{P}}
\newcommand{\IR}{\mathcal{IR}}

\usepackage{subcaption}
\begin{document}

\title{Copyright Infringement Risk Reduction via Chain-of-Thought and Task Instruction Prompting}

\author{Neeraj Sarna\inst{1}\thanks{email: nsarna@munichre.com} \and
Yuanyuan Li\inst{1}\thanks{email: yli@munichre.com} \and
Michael von Gablenz\inst{1}\thanks{email: mgablenz@munichre.com}}
\authorrunning{N. Sarna et al.}

\institute{Munich RE,\\
Keoniginstrasse 107, 80802,\\
Munich, Germany}

\date{}

\maketitle

\begin{abstract}
Large scale text-to-image generation models can memorize and reproduce their training dataset. Since the training dataset often contains copyrighted material, reproduction of training dataset poses a copyright infringement risk, which could result in legal liabilities and financial losses for both the AI user and the developer. The current works explores the potential of chain-of-thought and task instruction prompting in reducing copyrighted content generation. To this end, we present a formulation that combines these two techniques with two other copyright mitigation strategies: a) negative prompting, and b) prompt re-writing. We study the generated images in terms their similarity to a copyrighted image and their relevance of the user input. We present numerical experiments on a variety of models and provide insights on the effectiveness of the aforementioned techniques for varying model complexity. 
\end{abstract}

\section{Introduction}

Text-to-image generation models have shown remarkable capabilities in generating relevant and aesthetic images from user-input. However, these models can memorize their training dataset and reproduce a similar copy when prompted by the user \cite{carlini2023,Somepalli2023}. Since training sets often contain copyrighted content, memorization exposes both the AI user and the developer to copyright infringement risk, which often results in financial losses. Copyright infringement risk has already materialized for AI developer and is no more a hypothesis. Consider for instance, the case where artists sued Stability AI (Midjourney and DevianArt) claiming that the companies' AI image generators produce images that are strikingly similar to their artworks \cite{andersen2023}. Likewise, in August 2024, a U.S. district judge (Orrick) ruled that artists could proceed with copyright infringement claims, underscoring the ongoing legal uncertainties surrounding AI-generated content \cite{andersen_v_stability_2024}.

To promote a wider and safer adoption of generative models, copyright risk mitigation is crucial. The current work proposes such a risk mitigation strategy while focusing on \textit{copyrighted characters}---an example being Mario. As noted in \cite{fantastic_beasts}\cite{Sag2023,Henderson2023}, such characters are particularly challenging for two reasons. Firstly, these characters are a type of \textit{abstractly} protected content, which makes it "easier" for the generated output to infringe upon their copyright. Secondly, the training dataset might only contain characters in limited poses---standing, for example. However, a character generated in any other pose---dancing, for example---would still result in copyright infringement. Consequently, it is difficult to apply risk mitigation strategies designed for verbatim/exact memorization.

\textbf{Current work:} We focus on diffusion models and study the effectiveness of chain-of-thought (CoT) and task instruction (TI) prompting in mitigating copyright infringement risk. Our work is inspired from risk mitigation in Large Language Model (LLM) literature where both CoT and TI have been extensively studied to mitigate hallucination related risk \cite{WeiCoT2022} \cite{COV}\\\cite{hallucination_mitigation}
\cite{hallucination_review}. Motivated by this literature, we intend to apply CoT and TI to reduce copyright infringement. 

\begin{figure}
    \centering
    \includegraphics[scale = 0.3]{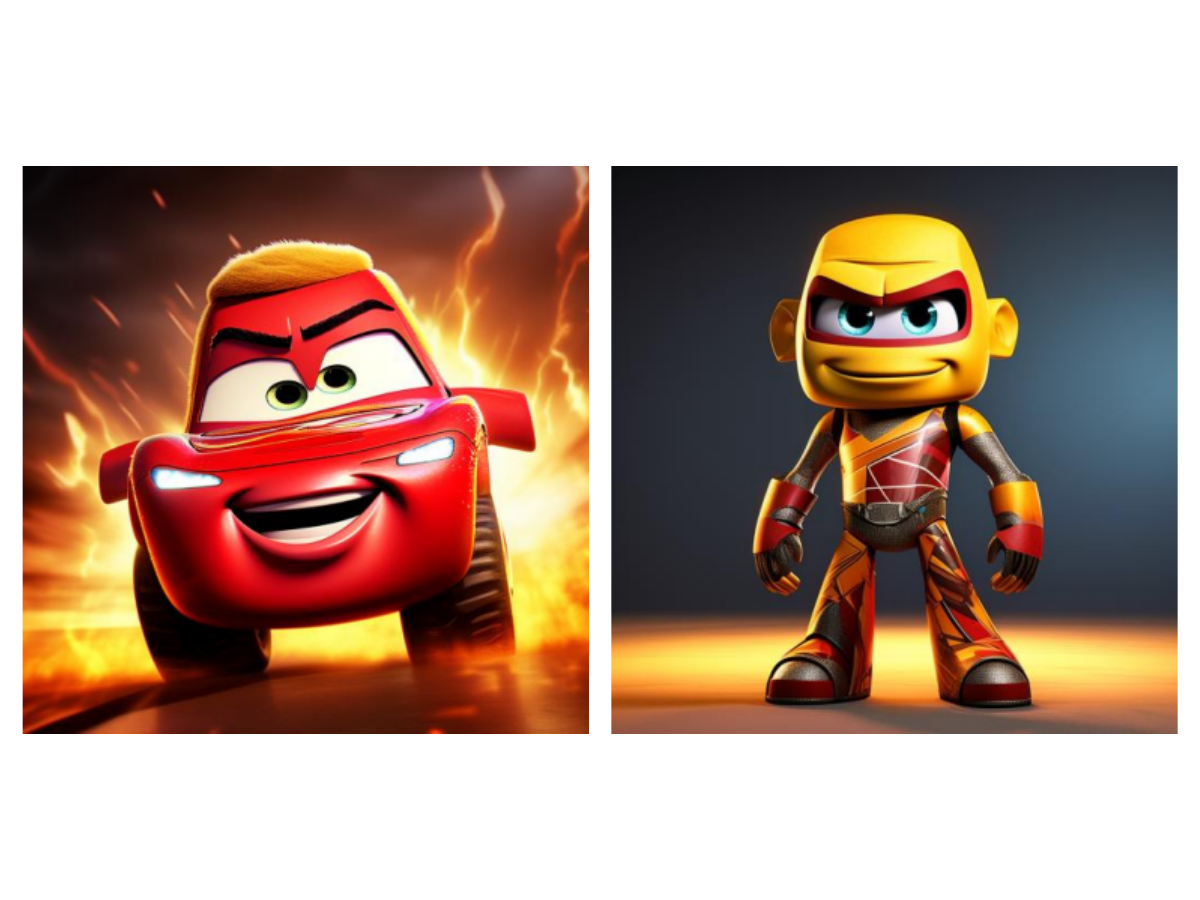}
    \caption{Results for PixArt model. Task Instruction prompting prevents copyright infringement. Prompt for: left) \textit{Generate an image of Lightining McQueen.}; and right) Task Instruction prompting that instructs the model to be creative and not to infringe any copyright (details in Section.~\ref{sec: prompt strategy})}
    \label{fig: example mcqueen}
\end{figure}

We present Fig-\ref{fig: example mcqueen} as an example. We generated an image of the character Lightining McQueen with the PixArt model. The model generates infringing content with the basic prompt. However, with TI prompting (details in Section-\ref{sec: prompt strategy}), we were able to generate a non-infringing image. In later sections, we explore this idea further by applying CoT and TI to a broader variety of models and characters. We also demonstrate that both CoT and TI can be combined with pre-existing copyright reduction strategy of negative prompting \cite{ho2021classifier}\cite{Ban2024} and prompt re-writing \cite{fantastic_beasts}\\ \cite{ramesh2021zero}, which, as we demonstrate, further improves the efficacy of these approaches. 

A copyright risk mitigation strategy must not only reduce the generation of copyrighted material but also produce a relevant image. Otherwise, irrespective of the user input, we can always output an image sampled from a random Gaussian, which would not infringe upon any copyrighted material but would be useless to the end user. We quantify this relevance using the cosine-similarity between CLIP embeddings, and study the ability of CoT and TI in maintaining this relevance. Furthermore, we investigate how infringement reduction can come at the cost of image relevance.

\textbf{Previous works: }Our work adds to the so-called \textit{black-box} risk mitigation strategies where the model user only has access only to the model output and not the internal states. A \textit{black-box} scenario better reflects---as compared to \textit{white-box}---the current state of GenAI models where the user access the model output via an API. To the best of our knowledge, in relation to copyright infringement reduction, CoT and TI has not been extensively studied in the literature especially when combined with negative prompting and prompt re-writing. We briefly summarize past works that align with ours. Authors in \cite{lena2025} explore CoT/TI in the context of memorization reduction while focusing on relatively smaller models, and applying CoT/TI without negative prompting or prompt re-writing. Another approach is
to use VLLMs to detect prompts that might generate copyrighted images. In case
such a prompt is detected, the diffusion process is guided away from copyrighted
outputs using negative prompting \cite{sonyAI_evaluation}. Authors in \cite{fantastic_beasts} study the effectiveness of prompt-rewriting, which has been deployed in commercial applications (like \cite{ramesh2021zero}). They conclude that although compared to the base prompt, prompt re-writing reduces the copyright infringement rate, best results are obtained when prompt re-writing is combined with negative prompting. This work inspired us to combine CoT/TI with negative prompting and prompt re-writing. Lastly, authors in \cite{Chiba2025}, use genericization to output the most general image thereby, avoiding the generation of copyrighted material. 

\section{Preliminaries}
This section presents the various prompting strategies used in the current work and our evaluation criteria for judging the quality of the generated image. We start with introducing the diffusion model. 

\subsection{Diffusion model}
A diffusion model proceeds in time steps. At time step $t$, conditioned on a user input prompt $\Pin$, it predicts the extent of denoising to be performed on a latent $z_t$. Under the classifier-free guidance approach, the denoising prediction for timestep $t-1$ is made using the equation \cite{ho2021classifier,Ban2024}
\begin{gather}
\mathcal{E}(z_t,\mathcal P_+,\mathcal P_-) := \omega\mathcal{E}_{\theta}(z_t,\mathcal{P}_{+}) + (1 - \omega) \mathcal{E}_{\theta}(z_t,\mathcal{P}_{-}), \label{eq: diffusion equation}
\end{gather}
where $\Pin_+$ and $\Pin_-$ capture the desired negative and positive characteristics of the image---the diffusion process is guided towards $\Pin_+$ and away from $\Pin_-$. The factor $\omega$ is a guidance scale, and a lower $\omega$ pushes the generation process further away from $\Pin_-$ . We set $\omega$ to the default of $7.5$ \cite{Ban2024}. We now present different choices for $\Pin_+$ and $\Pin_-$, which result in different prompting strategies.

\subsection{Strategies with $\Pin_- = \emptyset$}\label{sec: prompt strategy}
$\Pin_-$ is the so-called \textit{negative-prompt} and the generation process is guided away from it. For the strategies discussed in this section, we set
\begin{gather}
\Pin_- = \emptyset,
\end{gather}
where $\emptyset$ is an empty string. Note that the above choice of $\Pin_-$ results in the standard diffusion model without any steering away from unwanted characteristics \cite{ho2021classifier}. 

For the desirable characteristics encapsulated via $\Pin_+$, we build upon the base strategy. Example of a base prompt would be $\Pin^+_{Base}=$\textit{"Generate an image using the character description Mario."}, and same for other characters. In order to mold the different strategies into a general framework, we express this prompt more abstractly, $\Pin^+_{Base} =$\textit{Generate an image using the character description}+\textit{"Mario"}+$\emptyset$, where as before, $\emptyset$ is an empty string. In the following discussion, we would replace the empty string $\emptyset$ with elaborate descriptions thus generalizing the base strategy. 

We view $\{\textit{"Mario"}\}$ as a list of character description that we denote using $\chardes$. The variable $\chardes$ can also be a list of keywords describing the character Mario. We denote the character name using $\charname$ and a list of keywords describing the characters as $\keywords{\charname}$. Hence, depending on the prompting strategy, $\chardes$ is either $\{\charname\}$ or $k(\charname)$. We use the same characters as those proposed in \cite{fantastic_beasts}. Furthermore, we  choose the top-10 keywords using the language model (LM) based approach described in Section-3 of \cite{fantastic_beasts}---for completeness, Table.~\ref{table: char-key} presents some examples. The strings \textit{"Generate an image using the character"} and $\emptyset$ can be viewed as generation descriptions, which instruct the model on how to generate the image. We label them as $\gendes^{Base}_1$ and $\gendes^{Base}_2$, respectively, and collect them in the variable $\gendes_{Base}:=\{\gendes^{Base}_1,\gendes^{Base}_2\}$.

Using this notation, we can represent the base prompt as 
\begin{gather}
    \Pin^+_{Base} = \prompt{\gendes_{Base}}{\{\charname\}} = \gendes^{Base}_1 + \charname + \gendes^{Base}_2. 
\end{gather}
We define $\prompt{\gendes}{\chardes}$ as our general formulation of a prompt that takes as input the generation description $\gendes$ and the character description $\chardes$. In the general case, as noted above, $\chardes$ could be a list with more than one elements. Hence, our general prompt is expressed as
\begin{gather}
    \prompt{\gendes}{\chardes} = \gendes_1 + \sum_i \chardes_i + \gendes_2, 
\end{gather}
We capture the different prompting strategies using this prompt formulation. 

Prompt re-writing involves removing the character name and replacing it by a description/keywords \cite{ramesh2021zero}. This provides
\begin{gather}
    \mathcal{P}^+_{Re}(\charname) := \prompt{\gendes_{Base}}{\keywords{\charname}}.
\end{gather}
An example of a re-written prompt, for Mario, would be \textit{Generate an image using the character description red hat, mustache, blue overalls and white gloves}. Similarity, re-written prompts for other character can be obtained by combing the character names and the keywords for different characters.
Note that the re-written prompt could be more descriptive with details related to the motion, position, etc. of the character \cite{fantastic_beasts,ramesh2021zero}. For simplicity, we pursue the above approach and leave the exploration of a more descriptive prompt as a part of our future work.  

\begin{table}[h!]
\centering
\begin{tabular}{|l|p{10cm}|}
\hline
\textbf{Target} & \textbf{Description} \\ \hline
Ariel & Red hair, mermaid, green tail, purple seashells, blue eyes, youthful, slender, animated, adventurous, curious \\ \hline
Astro Boy & Jet-powered boots, pointy hair, red boots, round eyes, blue shorts, robotic, iconic, youthful, futuristic, manga character \\ \hline
Batman & Cape, cowl, bat-symbol, utility belt, dark colors, muscular, boots, gloves, mask, cape \\ \hline
Black Panther & Vibranium suit, claws, mask, black, stealthy, agile, muscular, necklace, superhero, Wakanda \\ \hline
\end{tabular}
\caption{Examples of Character Descriptions from the LM approach proposed in \cite{fantastic_beasts}.}
\label{table: char-key}
\end{table}

\textbf{CoT and TI prompts:}
There is no unique formulation for CoT and TI prompts. We rely on a trial-and-error approach where we start with a first draft for both, formulate different version using an LLM and select the version with the most desirable infringement rates and relevance scores---Appendix.~\ref{app: prompt details} provides further details. The generation descriptions for both CoT and TI are as follows.

\begin{tcolorbox}[colback=blue!5,colframe=blue!60!black,title=CoT (Generation Descriptions)]
$\gendes^{CoT}_1 = $\textit{Follow these steps: Begin by imagining a novel interpretation using the character description.} $\gendes^{CoT}_2=$ \textit{Next, ensure the final image reflects the essence of the character while adding your own creative twist and originality. Lastly, carefully consider the generated image and ensure that it is legally distinct from the copyrighted image.}
\end{tcolorbox}

\begin{tcolorbox}[colback=blue!5,colframe=blue!60!black,title=TI (Generation Descriptions)]
$\gendes^{TI}_1=$ \textit{Produce a unique character using the character description.} $\gendes^{TI}_2=$ \textit{but transformed into an entirely original creation. Use inventive shapes, different colors, and fantastical clothing to keep the design imaginative and legally distinct from copyrighted material.}
\end{tcolorbox}

With these descriptions, we arrive at the following prompting combinations. Choosing $\chardes$ as the character name $\{\charname\}$, we combine CoT and TI with the base prompt:
    \begin{gather}
        \mathcal P^+_{Base + CoT}:=\prompt{\gendes_{CoT}}{\{\charname\}},\quad \mathcal P^+_{Base + TI}:=\prompt{\gendes_{TI}}{\{\charname\}}.
    \end{gather}
Likewise, choosing $\chardes$ as the keywords $\keywords{\charname}$, we combine CoT and TI with re-written prompts:
    \begin{gather}
        \mathcal P^+_{Re+CoT}:=\prompt{\gendes_{CoT}}{\keywords{\charname}},\quad \mathcal P^+_{Re+TI}:=\prompt{\gendes_{TI}}{\keywords{\charname}}.
    \end{gather}

\subsection{Strategies with $\Pin_- \not = \emptyset$}
For the different strategies in this section, we set
\begin{gather}
\Pin_- = \charname,
\end{gather}
thereby, actively steering the generation process in (\ref{eq: diffusion equation}) away from the copyrighted character. Note that one could further extend $\Pin_-$ by including the character descriptions \cite{fantastic_beasts}. For simplicity, we do not pursue this extension.

The choice for $\Pin_+$ remains the same discussed above for $\Pin_- = \emptyset$. For completeness, we recall them here. When combining Base and Re with negative prompting (Neg), we have
\begin{gather}
 \Pin^+_{Neg + Base} = \Pin^+_{Base}, \quad \Pin^+_{Neg + Re} = \Pin^+_{Re}. 
\end{gather}
Adding CoT/TI to the above, we find
\begin{gather}
\Pin^+_{Neg+Base+CoT} = \Pin^+_{Base+CoT},\quad \Pin^+_{Neg+Base+TI} = \Pin^+_{Base+TI}, 
\end{gather}
and 
\begin{gather}
 \Pin^+_{Neg+Re+CoT} = \Pin^+_{Re+CoT}, \quad \Pin^+_{Neg+Re+TI} = \Pin^+_{Re+TI}. 
\end{gather}

\subsection{Evaluation Criteria}\label{sec: evaluation}
We evaluate an image on two criteria: i) Whether it results in copyright infringement? and ii) Given the prompt, how relevant is the image? The following discussion provides metrics to quantify these two attributes.

\textbf{Copyright infringement rate: }Let $\mathcal I(\charname)$ denote an indicator function which equals $1$ when the image corresponding to $\charname$ results in copyright infringement and $0$ otherwise. The copyright infringement rate then reads 
\begin{gather}
    \mathrm{InfRate} := \frac{1}{\#\mathcal S}\sum_{\charname\in \mathcal S}\mathcal I(\charname),
\end{gather}
where $\mathcal S$ is a set of all the considered characters and $\#\mathcal{S}$ is its cardinality. To determine $\mathcal I(C)$---same as \cite{sonyAI_evaluation}---we use human evaluation. We ensure that all the human evaluators are aware of all the characters and their original images in different poses. We acknowledge that unlike a VLLM-based or a distance metric based (L2/LPIPS) approach, human evaluation doesn't scale as the image set size grows \cite{Xu2025,fantastic_beasts}. Nonetheless, it offers a few benefits. Firstly, copyright detection based upon VLLMs is known to suffer from false positives, which could make a comparison across different infringement mitigation methodologies difficult \cite{Xu2025}. Secondly, a distance metric based approach leads to false negative even for some of the obvious infringement cases \cite{sonyAI_evaluation}. Lastly, a human evaluation better reflects the current scenario where a human judge determines copyright infringement. 

\textbf{Image relevance:} To capture the intent of the model user, similar to\\ \cite{fantastic_beasts}, we use the character keywords $\keywords{\charname}$. We use a CLIP encoder $E$ to encode a concatenation of these keywords, represented as $E(\sum \keywords{\charname})$. Likewise, we encode the generated image $X(\charname)$ via $E(X(\charname))$. Taking the cosine-similarity between the two encoding, we get the relevance score for a character $\charname$ and its average over the different characters
\begin{gather}
    \mathrm{Rel}(\charname) := \frac{\left\langle E(\sum \keywords{\charname},E(X(\charname))\right\rangle}{\| E(\sum \keywords{\charname}\|_2\|E(X(\charname))\|_2},\quad \mathrm{AvgRel} := \frac{1}{\#\mathcal S}\sum_{\charname\in\mathcal S}\mathrm{Rel}(\charname)
\end{gather}
where $\langle\cdot,\cdot\rangle$ represents the $L^2$ inner-product, and $\mathcal S$ is a set of all the characters considered in this work.

\section{Experimental Results}

\textbf{Models and dataset:} We consider the same $50$ comic characters as those considered in \cite{fantastic_beasts}. We consider three different models of varying complexity: Stable diffusion-2 (SD2), Stable diffusion-3 (SD3), and Pix-Art. We refer to \cite{rombach2022,SD3,PixArt} for further details related to these models. We do not perform any further fine-tuning on the models. For each of the models, we use $30$ inference steps (which struck a good balance between computational cost and image quality) and a guidance scale ($\omega$ in (\ref{eq: diffusion equation})) of $7.5$.

\textbf{Copyright infringement rates:} Table.~\ref{tab:infringe rates} presents the various infringement rates. For most cases, our results indicate that substantial $\IR$ reduction can be obtained by adding CoT and TI on top of Base, Re, Neg+Base and Neg+Re prompts. We elaborate further with the following observations.

\textit{For the SD2 model:} CoT and TI is the most effective in reducing the $\IR$. Consider Base + TI which, compared to Base, has a $85\%$ lower $\IR$. Even when comparing to Re, it has a $50\%$ lower $\IR$. Recall that Base+TI explicitly mentions the character names whereas Re does not. Hence, explicit character reference in a prompt doesn't necessarily results in infringement. Using TI style phrases in a prompt can help avoid infringement even for seemingly \textit{riskier} prompts.  
Drastic drops in $\IR$ are observed for negative prompting. Adding CoT/TI to both Neg+Base and Neg+Re results in a zero $\IR$---similar observations hold for the CoT prompt. 

\textit{For the SD3 model:} we observe a $8\%$ drop in $\IR$ when adding TI prompt to Base and Re prompts, which is lower compared to what we observed for the SD2 model. This indicates that larger models might have higher memorization, making it difficult (solely with prompt engineering) to avoid infringement for \textit{riskier} prompts which explicitly use the character name. Nonetheless, both TI and CoT offer way lower $\IR$ when combined with negative prompting. Adding TI to Neg+Base and Neg+Re lowers the $\IR$ by $50\%$ and $60\%$, respectively. Likewise, adding CoT to Neg+Base and Neg+Re lowers the $\IR$ by $40\%$ and $80\%$, respectively. At present it is unclear why no reduction in $\IR$ is observed when CoT and TI are added to Re. 

\textit{For the PixArt model:} Results for PixArt are very similar to the SD3 model. We observe a drop of almost $10\%$ in $\IR$ when TI are added to Base. Adding CoT/TI to negative prompting provides the most benefits. Adding TI to Neg+Base and Neg+Re lowers the $\IR$ by $60\%$. Likewise, adding CoT to Neg+Base and Neg+Re lowers the $\IR$ by $40\%$ and $80\%$, respectively. 

\textit{Best performing strategy:} For SD2 and PixArt, Neg+Re combined with TI results in the least $\IR$. For SD3, lowest $\IR$ is obtained when CoT is combined with Neg+Re. Note however that, other than SD2, the $\IR$ is still not zero for both SD3 and PixArt. Nevertheless, when compared to Base, the $\IR$ reduces by $98\%$ when Neg, Re and TI are combined. 

\begin{figure}
    \centering
    \begin{subfigure}{0.55\textwidth}
    \centering
            \includegraphics[width=0.9\linewidth]{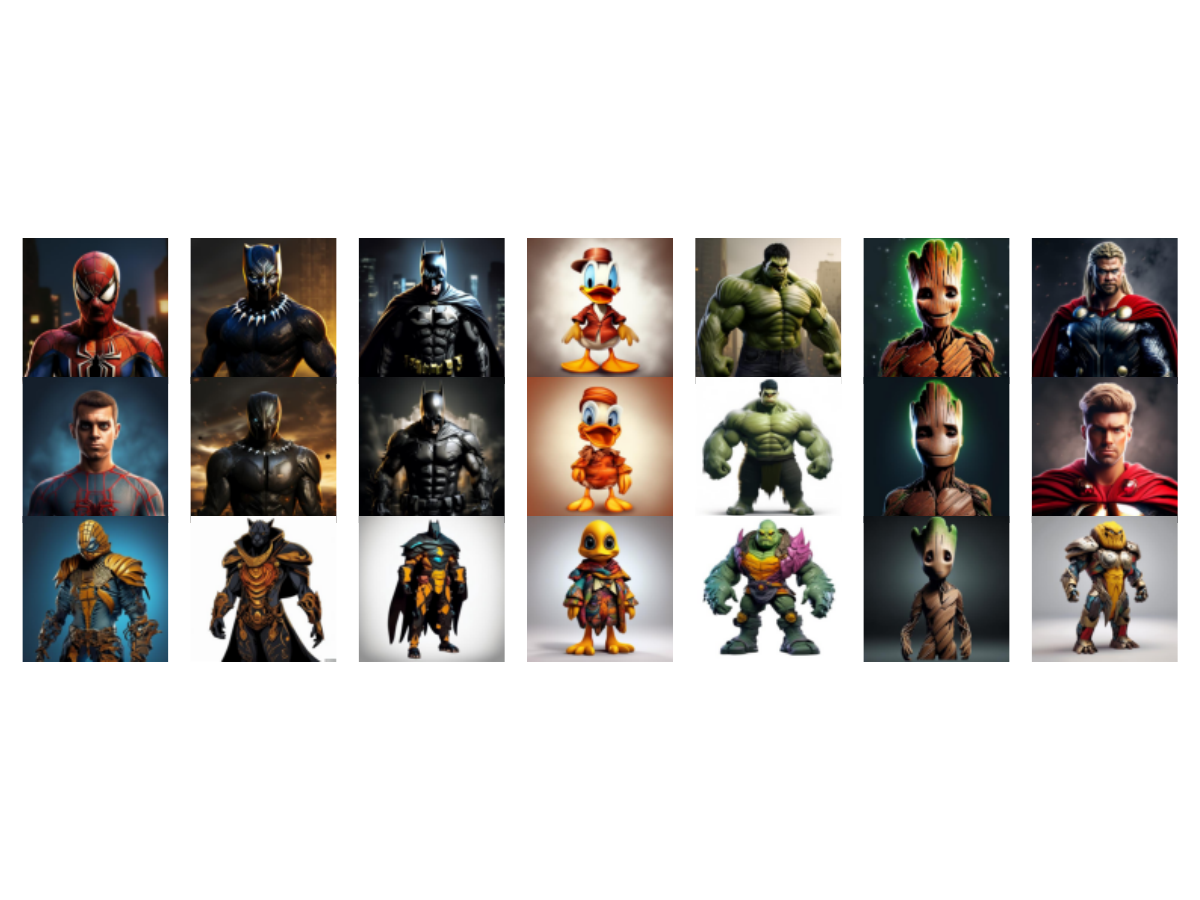}
    \caption{Results for PixArt. Top to bottom: Base, Neg+Base and Neg+Base+TI. }
    \label{fig: samples PixArt}
        \end{subfigure}
    \begin{subfigure}{0.4\textwidth}
    \centering
    \includegraphics[width=0.8\linewidth]{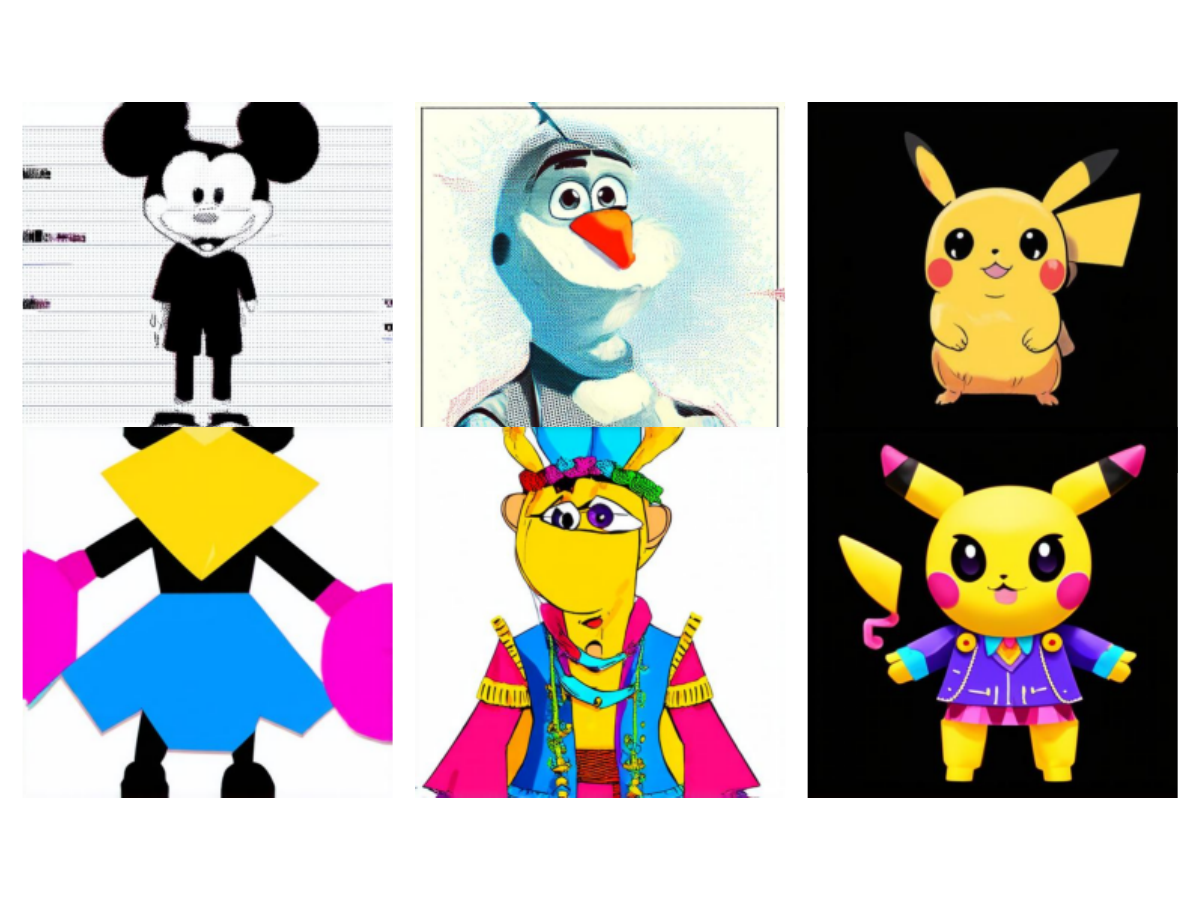}
    \caption{Results for SD3. Comparison for Neg+Base+CoT (top) vs. Neg+Base+TI (bottom).}
    \label{fig: cot vs task}
    \end{subfigure}
    \caption{a) Results for PixArt, depicting how addition of TI can help avoid infringement. b) Results for SD3, depicting a comparison between CoT and TI.}
\end{figure}

\textit{Samples from PixArt:} \ref{fig: samples PixArt} compares samples from Base, Neg+Base and Neg+Base+TI. For the base strategy, all the generations result in infringement. Other than the character Thor (last column), for Neg+Base, all the other generations result in infringement. Results improve substantially when TI is added to Neg+Base. Apart from Groot (second column from the right), none of the other characters result in infringement. We observe that the clothing of characters is unique and imaginative, which we attribute to the generation description of the TI prompt. Take Spider-Man for example, where a change in clothing make the character look very distinct from the copyrighted character---similar observation holds for Black Panther and Donal Duck. The character Groot is heavily memorized and for all three strategies, results in infringing content. 

\textit{CoT vs. TI:} In some scenarios, TI offers a drastically lower $\IR$ compared to CoT. Consider SD2, where Base+TI has a $75\%$ lower $\IR$ compared to Base+CoT. Similarly for SD3, Neg+Base+TI has $40\%$ lower $\IR$ compared to Neg+Base+CoT. We hypothesize that this is a consequence of the phrase \textit{Use inventive shapes, different colors, and fantastical clothing} in the TI prompt---samples similar to Fig.~\ref{fig: cot vs task} corroborate our hypothesis. For some characters, clothing, shapes and colors are crucial characteristics. When a generative model focuses on the aforementioned phrase and changes these characteristics, infringement could be avoided. Consider Olaf (middle column), whose change in clothing and color helps avoid infringement. Similarly, a rudimentary change in the shape of Mickey Mouse's face and clothing avoids infringement. However, change in clothing and color doesn't necessarily provide desirable results. Consider Pikachu, which, despite the change in clothing and some colors, still infringes copyright. Note that including the aforementioned phrase in CoT lead to similar results but to respect the original prompt generation process and avoid bias, we did not alter our CoT prompt. 

\begin{figure}[htbp!] 
    \centering
    \begin{subfigure}{0.45\textwidth}
            \centering
    \includegraphics[width=1\linewidth]{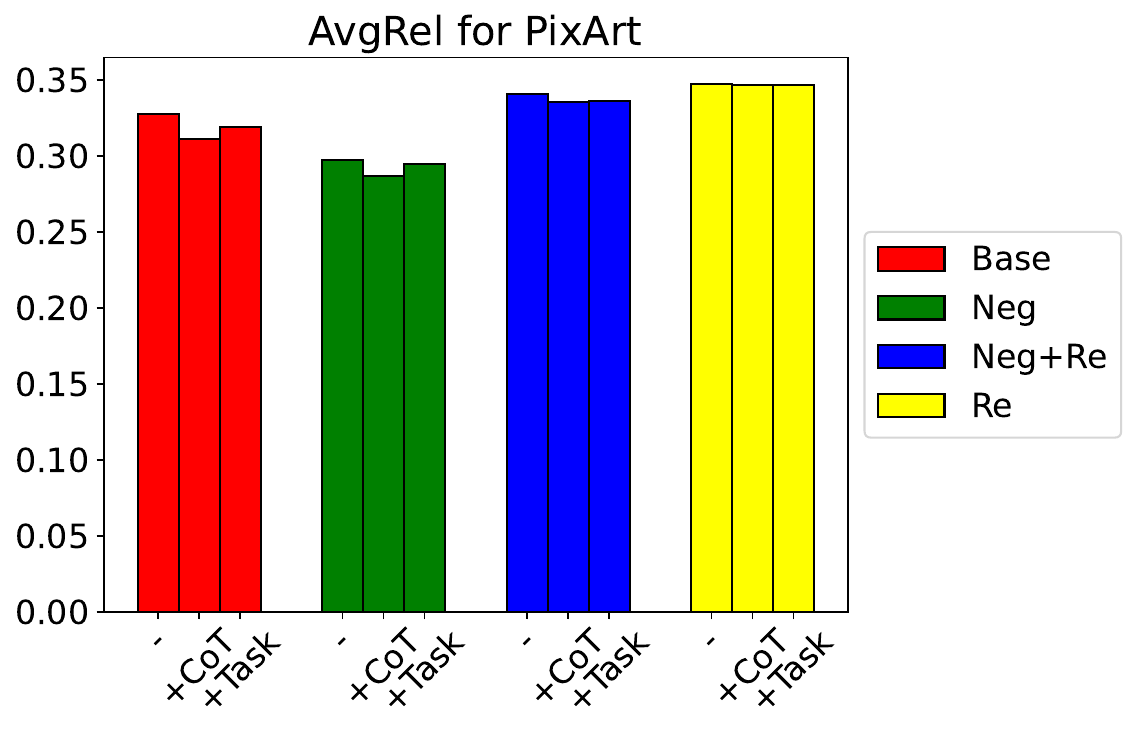}
    \caption{$\mathrm{AvgRel}$ for PixArt.}
    \label{fig: rel PixArt}
    \end{subfigure}
    \hfill
    \begin{subfigure}{0.45\textwidth}
    \centering
    \includegraphics[width=1\linewidth]{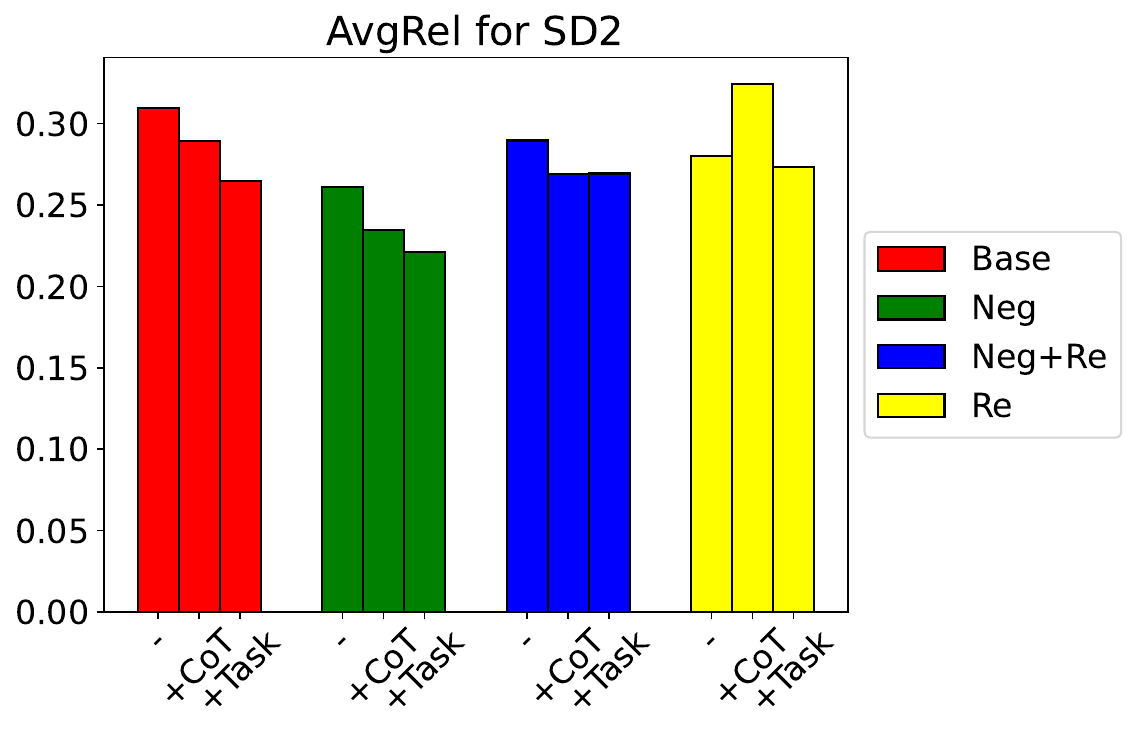}
    \caption{$\mathrm{AvgRel}$ for SD2.}
    \label{fig: rel SD2}
    \end{subfigure}
    \hfill
    \begin{subfigure}{0.45\textwidth}
    \centering
    \includegraphics[width=1\linewidth]{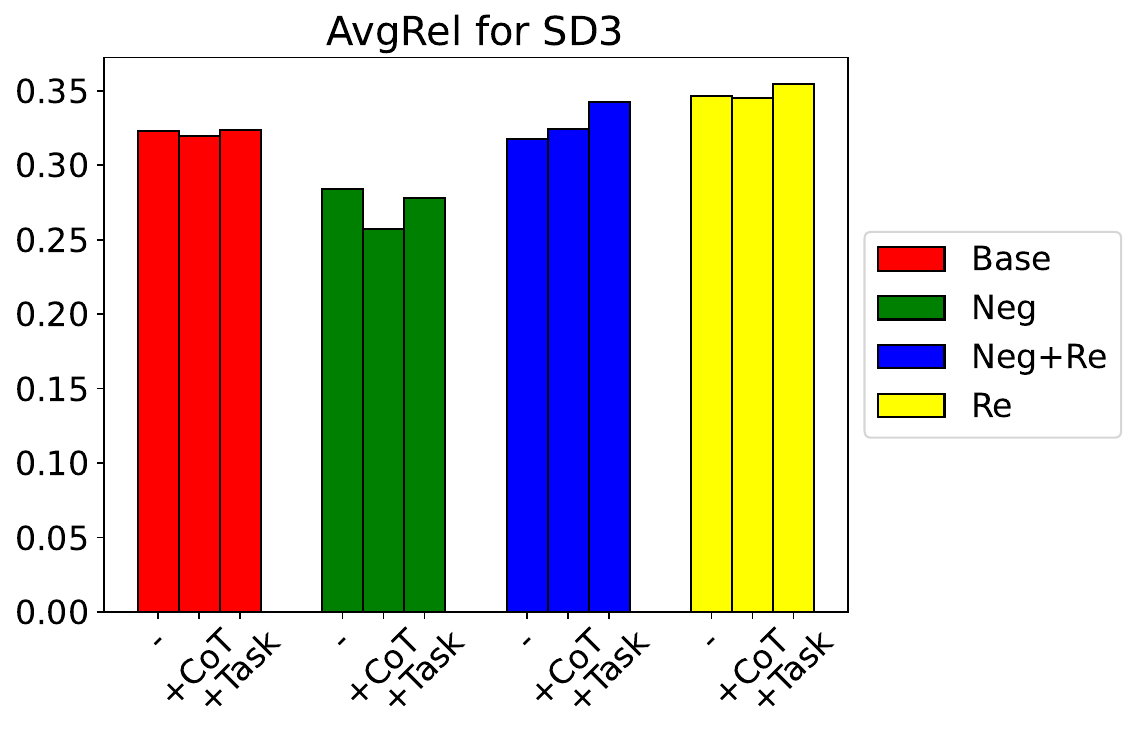}
    \caption{$\mathrm{AvgRel}$ for SD2.}
    \label{fig: rel SD3}
    \end{subfigure}
    \caption{$\mathrm{AvgRel}(\uparrow)$ for different models.}
    \label{fig: avgrel}
\end{figure}

\textbf{Relevance scores:} Fig-\ref{fig: avgrel} presents a bar-plot of different relevance scores. We make the following observations. Firstly, introducing CoT/TI to the various strategies has minor impact on the relevance scores. The largest drop of around $10\%$ is observed for SD2 with \textit{Base+TI} when compared to \textit{Base}. For all other cases, the difference is in the range of $0-5\%$. For SD2, AvgRel increases for Re+CoT when compared to Re. Secondly, $\mathrm{AvgRel}$ is slightly higher for strategies with prompt re-writing. This is as expected because we encapsulate the user intent using the character keywords instead of the character names---recall the discussion in Section.~\ref{sec: evaluation}. The re-written prompts explicitly contain these keywords, resulting in higher relevance scores. Lastly, introducing negative prompting reduces relevance---we elaborate on this further using generated samples. This is expected because we use the character name as the negative prompt. This \textit{strongly} guides the generation away from the character description, which is closely tied to the character name. 

\begin{figure}
    \centering
    \begin{subfigure}{0.55\textwidth}
    \centering
    \includegraphics[width=0.9\linewidth]{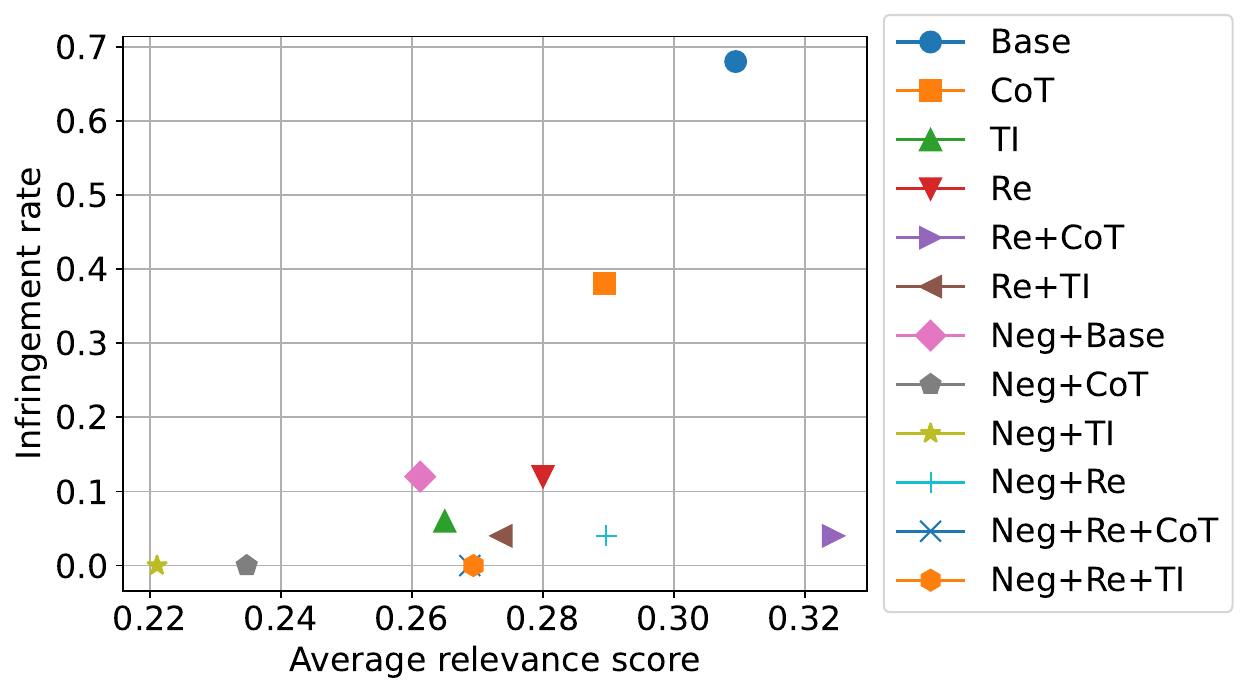}
    \caption{Results for SD2.}
    \label{fig: rel and ip}
        \end{subfigure}
    \begin{subfigure}{0.4\textwidth}
    \centering
    \includegraphics[width=0.9\linewidth]{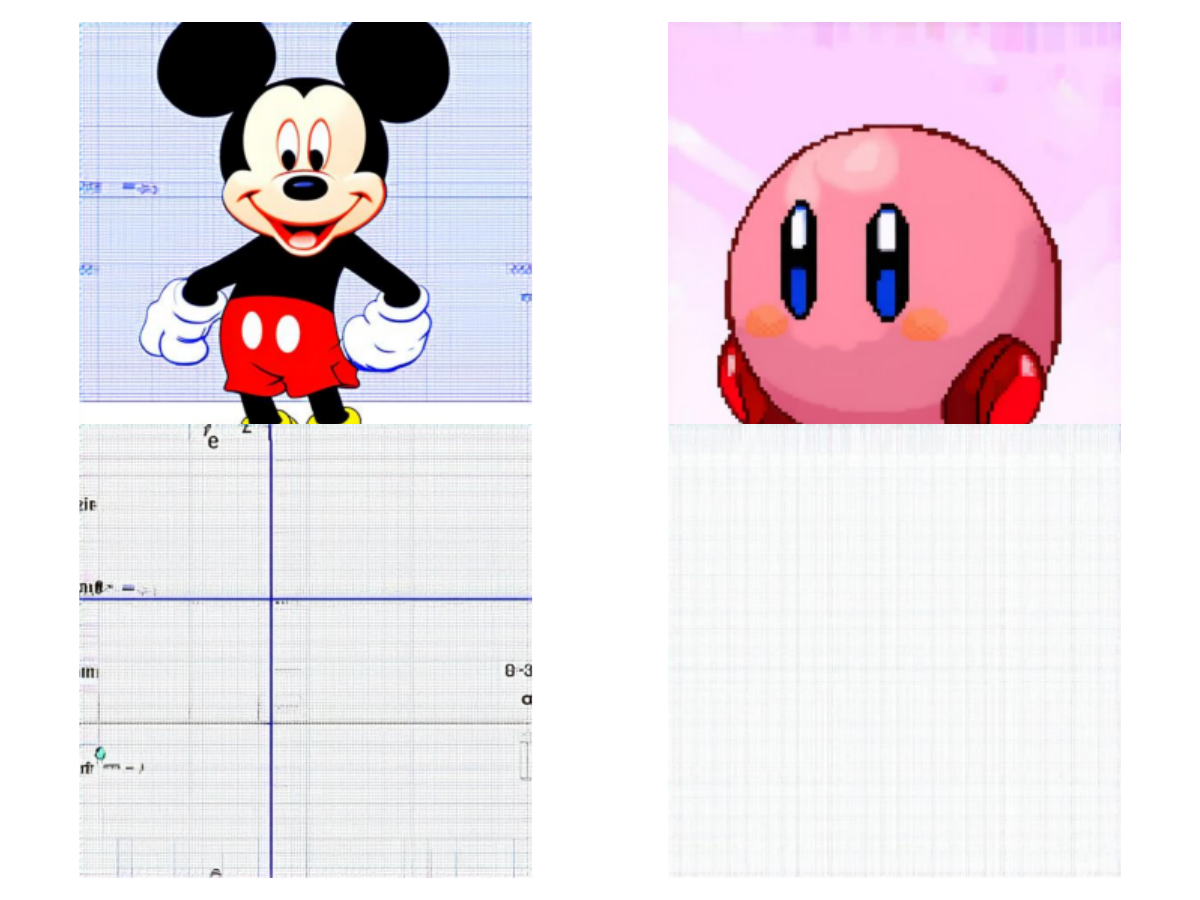}
    \caption{Results for SD3. Comparison between Neg+Base (bottom) and Base (top).}
    \label{fig: rel loss}
    \end{subfigure}
    \caption{a) Results for SD2 depicting how relevance reduces as infringement rates is reduced. b) Results for SD3 depicting how negative prompting can substantially decrease relevance for some characters.}
\end{figure}

\textbf{Relevance score vs. infringement rate:} Fig.~\ref{fig: rel and ip} plots the relevance scores against infringement rates; similar results were observed for other models and are not shown for brevity. We observe that decreasing the infringement rate comes at the cost of relevance. Consider Neg+Base for instance, which, compared to Base, has $90\%$ lower $\IR$ but also a $15\%$ lower relevance score. Observe that CoT+Re is an outlier---compared to Base, it decreases $\IR$ with a minor loss in relevance. As explained above, we use character description to capture user intent. Since Re strategies explicitly use this description (instead of the character name), we expect higher relevance. As Table.~\ref{tab:infringe rates} demonstrates, Neg based strategies are effective in $\IR$ reduction. However, for some characters, this comes at a high cost of relevance. Fig-\ref{fig: rel loss} presents examples where using Neg produces a blank image containing only the background. This prevents copyright infringement but the image losses relevance entirely. 

\begin{table}[ht!]
\centering
\caption{Copyright infringement rates ($\IR\downarrow$). \textbf{Best} and \underline{second} best method.}
\label{tab:infringe rates}
\resizebox{\textwidth}{!}{
\begin{tabular}{|l|c|c|c|c|c|c|c|c|c|c|c|c|c|}
\hline
 & \multicolumn{3}{c|}{Base} & \multicolumn{3}{c|}{Re} & \multicolumn{3}{c|}{Neg+Base} & \multicolumn{3}{c|}{Neg+Re}\\
\hline
model & - & +CoT & +TI & - & +TI & +CoT & - & +CoT & +TI & - & +CoT & +TI \\
\hline
SD2    & 0.68 & 0.38 & 0.06 & 0.12 & \underline{0.04} & \underline{0.04} & 0.12 & \textbf{0.00} & \textbf{0.00} & \underline{0.04} & \textbf{0.00} & \textbf{0.00} \\
\hline
SD3    & 0.78 & 0.80 & 0.72 & 0.38 & 0.36 & 0.28 & 0.16 & 0.12 & 0.08 & 0.10 & \textbf{0.02} & \underline{0.04} \\
\hline
PixArt & 0.66 & 0.62 & 0.60 & 0.32 & 0.28 & 0.26 & 0.44 & 0.32 & 0.18 & 0.10 & \underline{0.06} & \textbf{0.04} \\
\hline
\end{tabular}
}
\end{table}

\begin{figure}
    \centering
    \begin{subfigure}{0.45\textwidth}
    \centering
            \includegraphics[width=0.9\linewidth]{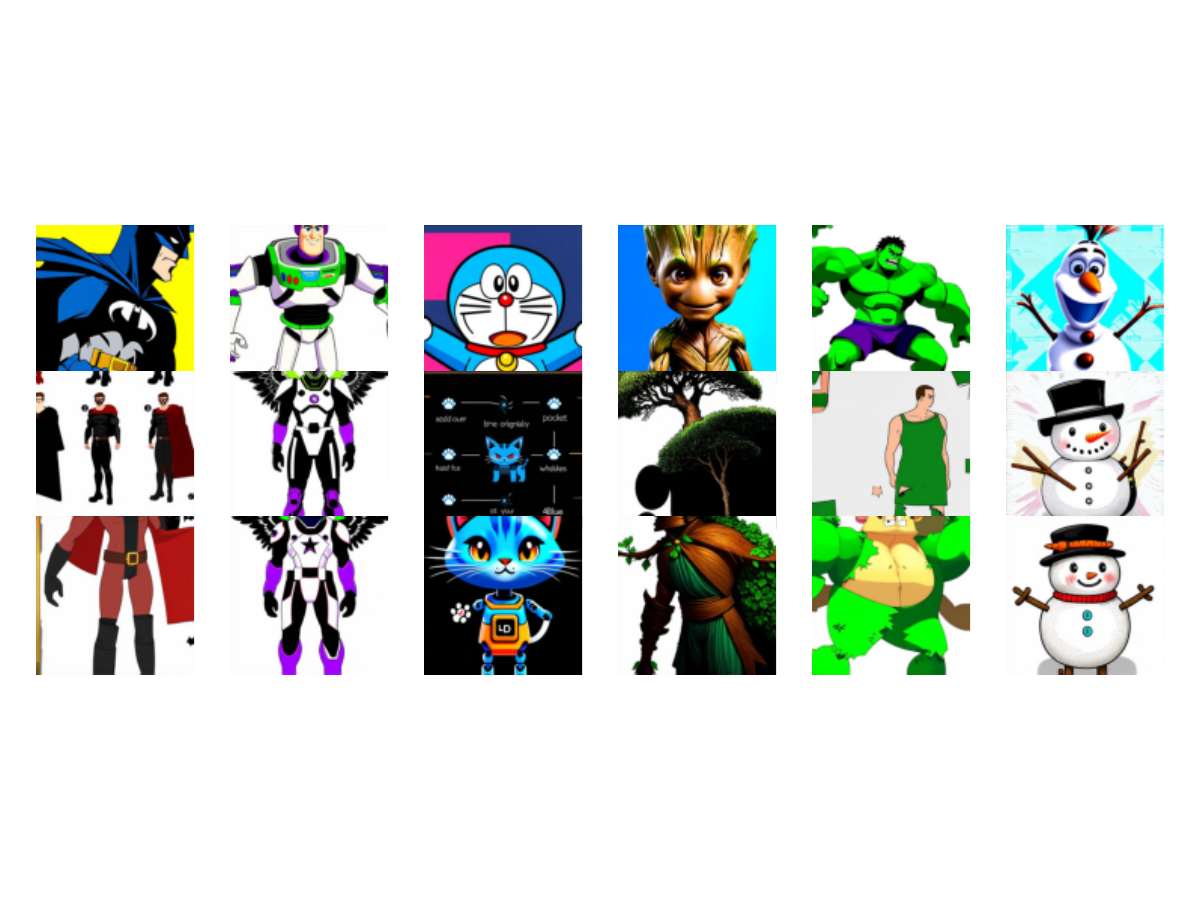}
    \caption{Results for SD3. Top to bottom: Base, Neg+Base+CoT and Neg+Base+TI. }
    \label{fig: worst_vs_best}
        \end{subfigure}
    \begin{subfigure}{0.45\textwidth}
    \centering
    \includegraphics[width=0.8\linewidth]{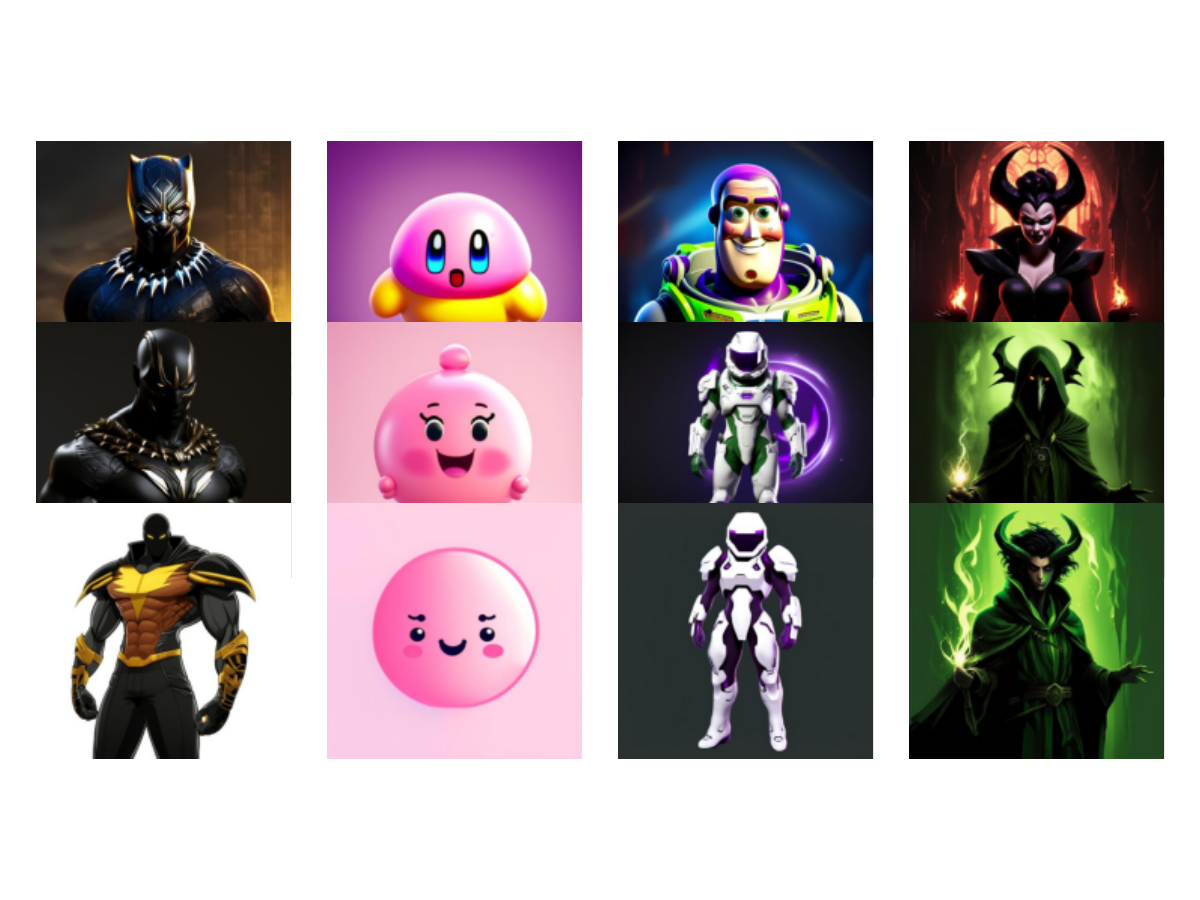}
    \caption{Comparison for: Base (top), Re+CoT (middle) and Neg+Re (bottom). Results for the SD3 model.}
    \label{fig: prompt_vs_neg}
    \end{subfigure}
    \caption{a) Results for SD3, comparison of the worst (Base) to the best (Neg+Base+CoT/TI) performing strategy. b) Results for SD3, comparison of prompt engineering to negative prompting.}
\end{figure}

\textbf{Comparison of best vs. worst strategy:} For SD3 and in terms of $\IR$ reduction, we compare the best (Neg+Re+CoT/TI) to the worst (Base) performing strategy. The results are presented in Fig.~\ref{fig: worst_vs_best}. Visually, the results from Neg+Re+CoT/TI maintain relevance while avoiding copyright infringement. Consider Olaf (first column from right), where the copyright is avoided by outputting a generic looking snowman---similar observation holds for Buzz Lightyear (second column from left) where a generic astraunaut with wings is generated. 

\textbf{Comparison of prompt engineering to negative prompting:} We compare prompt engineering (like Re+CoT) to negative prompting (Neg+Re+CoT). The former doesn't change the underlying distribution from which the image is sample. The latter, however, changes the underlying distribution by inversely scaling by the density of the unwanted characteristics---see \cite{ho2021classifier} for details. Furthermore, from a user perspective, prompt engineering is simpler to grasp. Unlike negative prompting, it doesn't require an understanding of diffusion-based generation process. Fig.~\ref{fig: prompt_vs_neg} presents a few examples comparing these two strategies. We observe that for some characters, prompt engineering provides similar results to negative prompting, while maintaining high relevance to the use input. 

\section{Conclusions}
While focusing on copyrighted characters, we studied the effectiveness of chain-of-thought (CoT) and task instruction (TI) prompting for reducing generation of copyrighted images in text-to-image models. Furthemore, we proposed a generic formulation that combines both these strategies with negative prompting and prompt re-writting. We performed numerical experiments on models with varying complexity. For a relatively smaller model (Stable Diffusion-2), by combining CoT/TI with negative prompting and prompt re-writing, we observed zero copyright infringement. Larger models, however, showcased a higher potential. Nonetheless, combining negative prompting with prompt re-writing and CoT/TI, we could reduce the copyright infringement rate from $70-80\%$ to $2-4\%$. Via exploiting CLIP embedding, we also studied the relevance of the generated images to the user's intent. We observed that adding CoT/TI to a prompting strategy has minor effect on the relevance of generations.

{\bibliographystyle{apalike}
\bibliography{AMS}
}

\appendix

\section{Details for prompting strategies} \label{app: prompt details}
\textbf{CoT/TI prompts ablation study:} We present our trial-and-error process to finalize the CoT/TI prompts used in the main text. For brevity, we outline the process for TI and using Mario as an example---the same process holds for CoT. We start with the prompt: $\Pin_1=$ \textit{Produce a unique character using the character description Mario but transformed into a original creation. Use novel ideas to keep the design imaginative and legally distinct from copyrighted material.}

Using this draft, we prompted GPT-5 to generate 5 different variants of the above prompt. The prompt we used for GPT-5 is as follows: \textit{You need to generate prompts for a text-to-image diffusion model such that copyright infringement is prevented. For the character Mario, here is an example for such a prompt:} $\{\Pin_1\}$. \textit{Propose 5 different alternatives to this prompt. Each of these 5 alternatives must follow the structure: String1 + character name + String2.}

In total, we generated 5 different TI prompts using the above strategy. For brevity, we present two examples: (i) Create a visually distinctive, highly creative, and non-copyright-infringing character using the character description Mario. Focus on originality and incorporate entirely novel visual elements. Avoid using recognizable characters, logos, or copyrighted designs. Ensure the image is imaginative and unique; (ii) Create an imaginative and entirely original character using the character description Mario but with a fresh design that avoids any copyrighted traits. Focus on inventing new shapes, colors, attire, and accessories, ensuring the result feels unique and not a copy of existing characters.

\begin{figure}[htbp!] 
    \centering
    \begin{subfigure}{0.45\textwidth}
    \centering
    \includegraphics[width=1\linewidth]{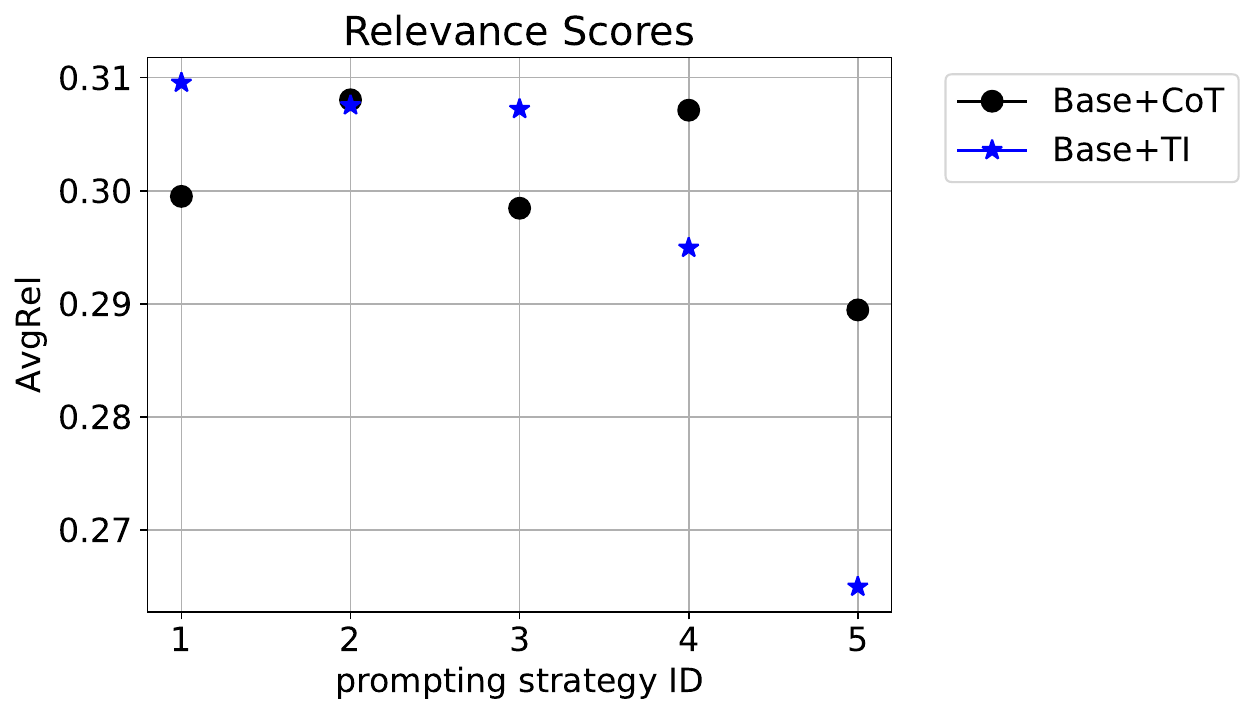}
    \end{subfigure}
    \hfill
    \begin{subfigure}{0.45\textwidth}
    \centering
    \includegraphics[width=1\linewidth]{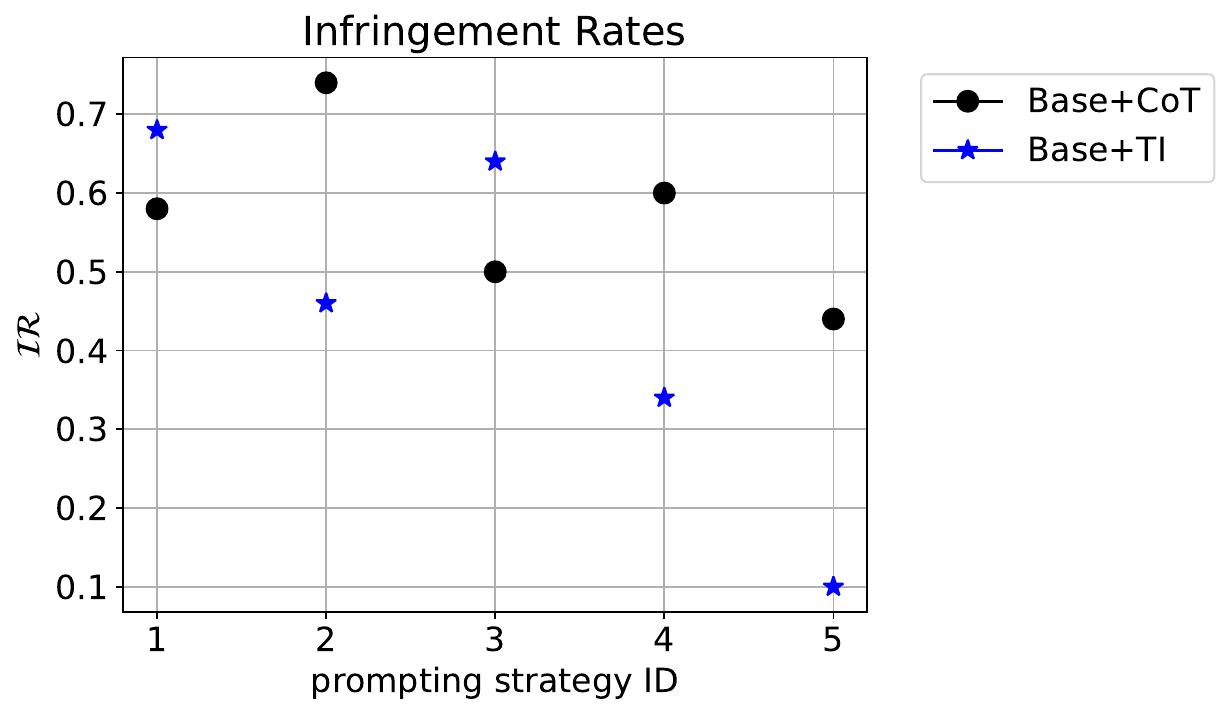}
    \end{subfigure}
    
    \caption{(a) $\mathrm{AvgRel}$ ($\uparrow$) and (b) $\IR$ ($\downarrow$) under prompt ablation.}
    \label{fig: ablation results}
\end{figure}

For the SD2 model, infringement rates $\IR$ and average relevance scores $\mathrm{AvgRel}$ for different strategies are shown in Fig.~\ref{fig: ablation results}. Strategy-5, used in the main text, strikes the best balance between $\IR$ reduction and relevance. 

\end{document}